# Video Event Recognition and Anomaly Detection by Combining Gaussian Process and Hierarchical Dirichlet Process Models

Michael Ying Yang[1], Wentong Liao[2], Yanpeng Cao[3] and Bodo Rosenhahn[2]

**Abstract**

In this paper, we present an unsupervised learning framework for analyzing activities and interactions in surveillance videos. In our framework, three levels of video events are connected by Hierarchical Dirichlet Process (HDP) model: low-level visual features, simple atomic activities, and multi-agent interactions. Atomic activities are represented as distribution of low-level features, while complicated interactions are represented as distribution of atomic activities. This learning process is unsupervised. Given a training video sequence, low-level visual features are extracted based on optic flow and then clustered into different atomic activities and video clips are clustered into different interactions. The HDP model automatically decide the number of clusters, i.e. the categories of atomic activities and interactions. Based on the learned atomic activities and interactions, a training dataset is generated to train the Gaussian Process (GP) classifier. Then the trained GP models work in newly captured video to classify interactions and detect abnormal events in real time. Furthermore, the temporal dependencies between video events learned by HDP-Hidden Markov Models (HMM) are effectively integrated into GP classifier to enhance the accuracy of the classification in newly captured videos. Our framework couples the benefits of the generative model (HDP) with the discriminant model (GP). We provide detailed experiments showing that our framework enjoys favorable performance in video event classification in real-time in a crowded traffic scene.

## I. Introduction

High-level video event classification is an important issue in computer vision and have attracted great attention in recent years [1] due to their significant practical values such as security monitoring, traffic controlling, etc. Most existing approaches focused on recognition of an individual activity [2], or a collective activity [3] in clean backgrounds. This task remains challenging in a crowded public scene due to a large number of agents with different activities at the same time, and complicated interactions such as traffic flows at a busy junction. Moreover, a surveillance video captured from a crowded scene normally is low-quality.

[1]Scene Understanding Group, ITC Faculty, University of Twente, **michael.yang@utwente.nl**
[2]Institute for Information Processing, Leibniz University Hannover
[3]Zhejiang University

Discriminant models such as GP models and SVM are the most popular approaches to classify video event [4], [5], [6], [7] because of their advantage in terms of classification accuracy. However, they are supervised model and a training data set with manual label is necessary in advance. Besides, they are feature-based approaches. They have high requirement in the applicability and the preciseness of features to ensure their performance. The most widely used features include HOG feature, flow-based features, etc.

Generative models especially the topic models such as LDA [8] and HDP [9], [10] have achieved great progress in high-level video event recognition in the complex surveillance scenes. They effectively learn activities and interactions from non-labeled video by analyzing semantic relationships instead. However, they have serious limitations-consuming computation and work in batch. Besides, most existing methods neglect the temporal dependencies between activities and interactions [9].

Inspired by the power of generative and discriminative models, in this paper, we propose a method to combine the HDP models and the GP models to realize unsupervised video behavior classification in real-time in a complex and crowded traffic scene. The first step is unsupervised learning the activities using HDP models and traffic states using HDP-HMM, respectively. Based on their learning results, we construct feature vectors to represent activities and traffic states in a new way. A training set is then generated with these feature vectors to feed the GP models. In addition, the temporal dependencies between two states are integrated into our GP models to enhance classification accuracy.

The major contributions of this paper are following. First, we effectively combine unsupervised generative model HDP with supervised discriminant model GP, to realize unsupervised classification of video event. Second, we integrate transition information between two states with GP models to enhance the accuracy of classification. Third, we provide detailed experiments showing that our framework enjoys favorable performance in video event classification in real-time in a crowded traffic scene.

## II. RELATED WORK

Topic models have received increasing attention to analyze activity in surveillance video [8], [11], [10], [12], [9], [7]. However, [12], [9] are offline and batch procedures and temporal dependencies are neglected. [8], [11] used latent Dirichlet allocation (LDA) models to infer activities in a video, which requires predefined number of clusters. It is hard to give a proper number of possible activities that may occur in a video from a crowded scene. Besides, their models perform Gibbs sampling in each newly captured video clip to estimate the joint distribution. It is time consuming and especially inefficient in an online model.

GP models have been applied for human motion analysis and activities recognition [4], [13] because of its robustness and high accuracy in classification. However, GP models are supervised. They must be fed with manually labeled data set. On the other hand, GP models require proper features to model events such as the most widely used trajectories [14], [15]. However, tracking-based methods depend crucially on the performance of detection and tracking which is costly or even impossible in our complex and crowded scene. Li *et al.* [6] proposed to detect anomalies in crowds utilizing Gaussian process regression models, which adopts HOF features to describe motion patterns. But their work is unable to analyze individual activity and interaction occurring in the surveillance scene. Hu *et al.* [16] combined the HDP model with One-Class SVM by using Fisher kernel. Tang *et al.* [17] proposed an alternative method to combine features for complex event recognition. However, this method is unfeasible in a surveillance video because of the low-quality video and too many objects. The low-level visual features are much more applicable in this scene.

## III. VISUAL FEATURES REPRESENTATION

Our datasets are surveillance videos from complex and crowded traffic scenes and captured by a fixed camera. They contain a large quantity of activities and interactions. Many unavoidable problems such as occlusions, a variety object types, small size of objects challenge detection and tracking based methods. In such case, using the local motions as low-level features is a reliable way. Firstly, the optical flow vector for each pixel between each pair of successive frames is computed using [18]. A proper threshold is necessary to reduce noise: the intensity of a flow is greater than the threshold is deemed as reliable. Similar to [9], [10], [19] we spatially divide the camera scene into non-overlapping square cells of size $8 \times 8$ pixels to get rough position features. We average all the optical flow vectors in the cell and quantize it into one of the 8 directions (Fig. 4(c)) as a local motion feature. A low-level feature is defined as the position of the cell $(x, y)$ and its motion direction. The image size of the two QMUL datasets [8] is $360 * 288$, thus they have $12960$ words, while the MIT dataset [9] $(480 * 720)$ has $43200$ words. Each word is represented by an unique integer index. The input videos are uniformly segmented into non-overlapping clips for $75$ frames each (3 seconds) and each video clip is viewed as a document which is a bag of all visual words $\mathbf{w}_t$ occurring in the $t^{th}$ clip. The whole input video is a corpus.

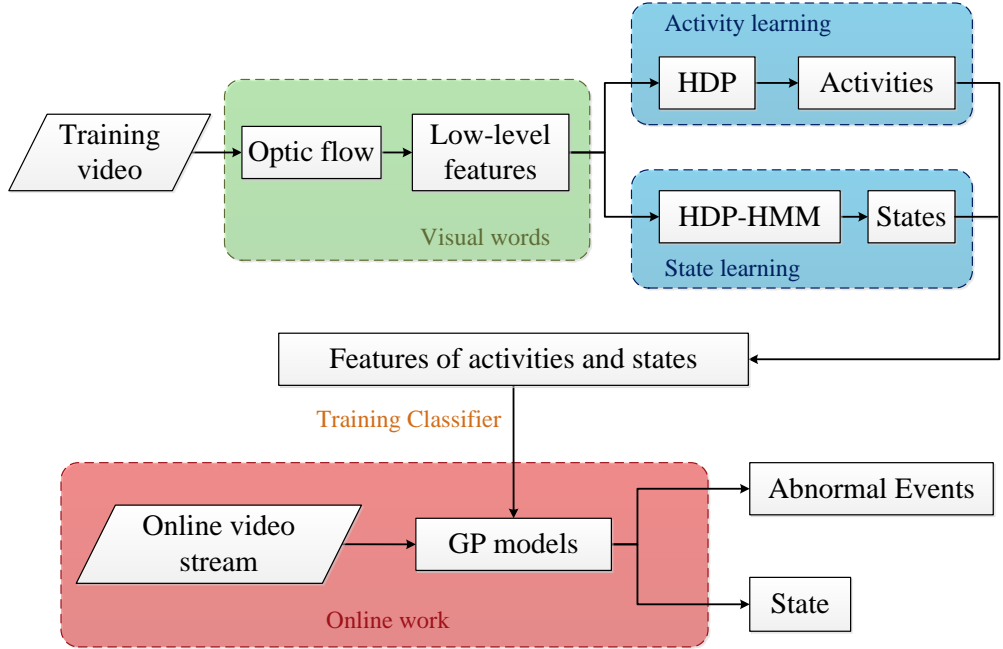

Fig. 1: An overview of our proposed framework. It's roughly divided into 3 parts. In the first part (in the green block), visual words are generated based on location in the image plane and direction to represent quantized motion information. Then, the HDP models learn the activity patterns in an unsupervised way (in the blue block). Finally, the learned patterns are used to train the GP models (in the red block) for our final goal of this work: activity recognition and anomaly detection.

## IV. MODEL

Our first task is to infer the typical activities and traffic states from given video. The low-level features are the exclusive motion information that can be directly observed from the input video. An activity is a mixture of local motions that frequently co-exist in the same clips (or documents).

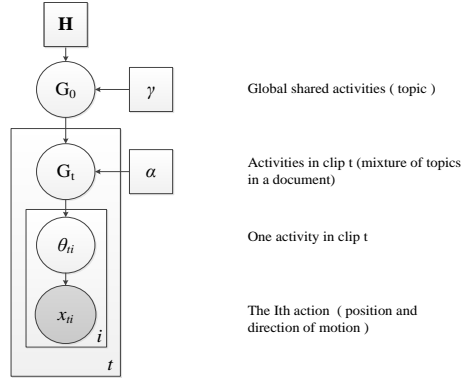

Fig. 2: A graphical representation of HDP model. It consists of two Dirichlet Processes. The first one is used to generate a global set of activities and the second one is used to sample a subset of activities from the global set for a clip. Finally, visual words are drawn from activities.

Thus, it is equivalent to infer topics in word-document analysis. Moreover, a traffic state is a combination of frequently co-occurring activities (i.e. interactions). This makes it possible to infer traffic states using topic model, too.

The HDP [20] is an unsupervised non-parametric hierarchical Bayesian topic model and was originally proposed for word-document analysis. It clusters the frequently co-occurring words within the same documents into the same topics. Furthermore, different from the other clustering topic models, such as LDA [21], HDP is able to automatically determine the number of clusters. The rest of this section will show how to use HDP model to infer typical activities and traffic states from the input video. Based on the output of HDP models, we propose a method to construct feature vectors to represent activities with visual words and traffic states with typical activities. Afterward these will be used to train classifier to recognize complicated traffic activities in surveillance video.

*A. Learning activities using HDP*

The possible activities are inferred by HDP whose standard graphical representation is shown in Fig. 2 [20]. The global random measure $G_0 = \{\theta_1...\theta_\infty\}$ is a global list of activities that is shared by all clips. Its distribution is a Dirichlet Process (DP) with concentration parameter $\lambda$ and Dirichlet prior $H$:

$$G_0|\gamma, H \sim DP(\gamma, H) \qquad (1)$$

$G_0$ can be expressed using the stick-breaking formulation [20]:

$$G_0 = \sum_{k=1}^{\infty} \pi_{0k} \delta_{\phi_k}, \tag{2}$$

$$\phi_k | \gamma \sim H, \tag{3}$$

$$\pi_k = \pi'_k \prod_{l=1}^{k-1} (1 - \pi'_l), \tag{4}$$

$$\pi'_k \sim Beta(1, \lambda), \tag{5}$$

where $\{\phi_k\}_k^{\infty} = 1$ are parameters of multinomial distributions over words in the codebook corresponding to activity $\theta_k$, i.e. the word probability vector and the sum of its entries equals 1. $\delta_{\phi_k}$ is the Delta function at point $\phi_k$. $\{\pi_k\}$ are random probability measures (mixtures over topics) and $\Sigma_{k=1}^{\infty} \pi_k = 1$. For convenience, the random probability measure of $\pi$ defined from (2) to (5) is abbreviated with $\pi_k \sim GEM(\gamma)$, where GEM stands for Griffiths-Engen-McCloskey distribution [22]. The multinomial distribution $\phi_k$ over words in the codebook is generated from $H$. Therefore, $H$ is interpreted as a distribution over multinomial distributions and thus can be defined as a Dirichlet distribution:

$$H = Dir(D_0), \tag{6}$$

$$\phi_k | \gamma \sim Dir(D_0). \tag{7}$$

$G_0$ is the prior distribution for the second DP. For each clip $t$, $G_t$ is a random measure which is drawn from the second DP with concentration parameter $\alpha$ and Dirichlet prior $G_0$:

$$G_t | \alpha, G_0 \sim DP(\alpha, G_0). \tag{8}$$

In our case $G_t$ describes the multinomial distribution of active topics in clip $t$, i.e. it is a subset of the global activities $G_0$. We express it using the stick-breaking representation again:

$$G_t = \sum_{k=1}^{\infty} \pi_{tk} \delta_{\phi_k}, \tag{9}$$

$$\phi_k | \alpha, G_0 \sim G_0, \tag{10}$$

$$\pi_{tk} = \pi'_{tk} \prod_{l=1}^{k-1} (1 - \pi'_{tl}), \tag{11}$$

$$\pi'_{tk} \sim Beta(1, \alpha). \tag{12}$$

For the $i^{th}$ word in document $t$, a topic $\theta_{ti}$ is first drawn from $G_t$ and then the word $x_{ti}$ is drawn from multinominal distribution $Multi(x_{ti}; \phi_{\theta_{ti}})$ (i.e. the multinominal distribution over words in codebook corresponding to topic $\theta_{ti}$). We notice that, different $G_t$ has the same $\phi_k$ as $G_0$, i.e. different clips share the same set of topics and statistical strength. We apply Gibbs sampling schemes to do inference under an HDP model, which is a generally applied method in topic model. Fig. 6 shows the learned typical activities by HDP models for QMUL Junction Dataset [8].

The hyper-parameters $\gamma$ and $\alpha$ are empirically predefined. They are priors on the concentration of the word distribution within topics. They influence the the number of activities in $G_0$ and $G_t$. The parameter $D_0$ for the Dirichlet distribution is also set empirically.

Although HDP models decide the number of topics automatically, some of the explored activities are unrepresentative. Because some very rare motion need to be explained by an individual activity. They could be noise or rare events. Such learned activities could lead to ambiguous or even misleading analysis of interactions. Therefore, the unrepresentative activities need to be removed. The total number of words that are assigned to activity $k$ is noticed as $n_k$ throughout the training video. The occurrence ratio of activity $k$ is computed as

$$r_k = \frac{n_k}{n_1 + \cdots + n_K}. \tag{13}$$

We rank $\{r_1, \cdots, r_K\}$ in decreasing order as $\{r'_1 \geq \cdots \geq r'_K\}$ and calculate the accumulated sum as

$$R'_j = \sum_{i=1}^{j} r'_i \tag{14}$$

The representative activity (topic) set is selected as

$$\theta_{typical} \triangleq \{\theta_j | R'_j \leq 0.99\}, \quad 1 \leq j \leq K, \tag{15}$$

### B. Learning states using HDP-HMM

A busy traffic junction is normally regulated by traffic lights: different traffic states occur sequentially and circulatory in a certain order. Hidden Markov model (HMM) [23] is an efficient method to explore the latent states and their transition information. HMM can be explained as a doubly stochastic Markov chain and is essentially a dynamic variant of a finite mixture model.

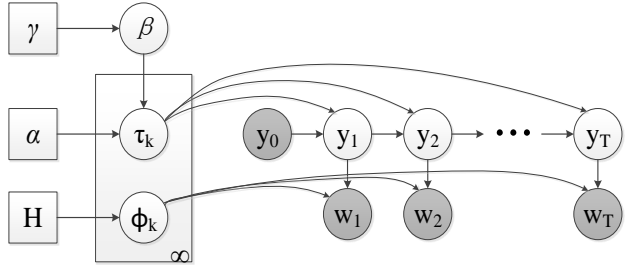

Fig. 3: A graphical representation of the HDP-HMM model.

[20] replaced the finite mixture with a Dirichlet process and proposed the HDP-HMM model which is illustrated in Fig. 3. Its stick-breaking formalism is:

$$\beta \sim GEM(\gamma), \tag{16}$$

$$\tau_k \sim DP(\alpha, \beta), \tag{17}$$

$$\phi_k \sim H \tag{18}$$

$$y_t | y_{t-1} \sim Multi(\tau_{y_{t-1}}), \tag{19}$$

$$\mathbf{x}_t | y_t = s_i \sim Multi(\phi_{s_i}). \tag{20}$$

where $y_t \in \mathbf{S} = \{s_1, \cdots, s_{N_s}\}$ is the state of the $t^{th}$ clip and $\mathbf{S}$ is the set of possible states and $N_s$ is the total number. $\mathbf{x}_t$ is the observation set (visual words). In this case, each vector $\tau_k = \{\tau_{kl}\}_{l=1\cdots L}$ is one row of the Markov chain's transition matrix from state $k$ to the other states and $L$ is the number of states. For a better illustration, we denote these transition matrix as $\mathbf{M} = \{m_{i,j}\}_{i,j=1\cdots L}$ throughout this paper. Given the state $y_t$, the observation $\mathbf{x}_t$ is drawn from the mixture component $\phi_{s_i}$ indexed by $y_t$. Gibbs sampling schemes are applied to do inference under this HDP-HMM. Fig. 7 shows the typical traffic states learned by HDP-HMM for QMUL Junction Dataset [8].

The same as the activity learning using HDP model, the traffic states learned by HDP-HMM also involve some unexpected results. The typical traffic states are selected in the similar way as described in Sec. IV-A.

## C. Representation of Activities and Video Clips

*Activity Representation:* Each activity $\theta_k$ is characterized by a multinominal distribution $\{\phi_k\}$ over the words in codebook. The probability of $i^{th}$ word in activity $\theta_k$ is denoted as $p_{kx_i}$ and $\mathbf{p}_{k\mathbf{x}} = \{p_{kx_i}\}_{i=1}^{N_x}$, $\Sigma_{i=1}^{N_x} p_{kx_i} = 1$ and $N_x$ is the size of codebook. Similar to the operation in Sec.IV-A which selects the representative activity, we also select the representative visual words to represent each activity in the same way: $\mathbf{p}_{k\mathbf{x}}$ is sorted in descending order $\mathbf{p}'_{k\mathbf{x}} = \{p'_{kx_1} \geq \cdots \geq p'_{kxN_x}\}$ and then the accumulated sum of probability is calculated as:

$$P'_{kj} = \sum_{i=1}^{j} p'_{kx_i} \tag{21}$$

those visual words which satisfy :

$$\mathbf{w}_{\theta_k} = \{x_j | P'_{kj} \leq 0.9\} \tag{22}$$

are chosen to represent activity $\theta_k$. It is the set of the most frequently co-occurring words in the same activity. The words falling into the rest $10\%$ are viewed as noise or rare motion. Fig. 4 shows an comparison between all possible co-occurring visual words and the selected representative words in the activity of vehicles driving downward.

*Video Clip Representation:* Feature vectors of activities from last step are variant in length because the number of representative words of different activities is unexpected. They are not suitable to be used to describe a video clip directly. We construct a feature vector to explain a clip using learned activities in a new way as follows.

$\mathbf{x}_t = \{x_{ti}\}_{i=1}^{N_t}$ denotes that there are $N_t$ the words present in clip $t$ totally. $\mathbf{x}_t$ is compared with each activity word set $\mathbf{w}_{\theta_k}$ and the percentage of intersection is calculated as:

$$p_{tk} = \frac{\mathbf{x}_t \cap \mathbf{w}_k^a}{N_t} \tag{23}$$

It explains the proportion of activity $\theta_k$ in this clip. The feature vector which explains what happens in this clip is represented as $\mathbf{c}_t = \{p_{t1}, \cdots, p_{tK}\}$, as shown in Fig. 7(e)-(h).

## V. TRAFFIC STATES CLASSIFICATION

In this section, we first discuss how to use GP models to classify traffic states in a newly screened video. Then we integrate the transition information learned by HDP-HMM with GP model to enhance the classification accuracy.

### A. Gaussian Process for Classification

The HDP-HMM has mined the main traffic states $\mathbf{S}$ from training video sequence and each training video clip is labeled with a state label $y_t \in \mathbf{S}$, where the subscript $t$ is the clip index. $\mathbf{c}_t\}$

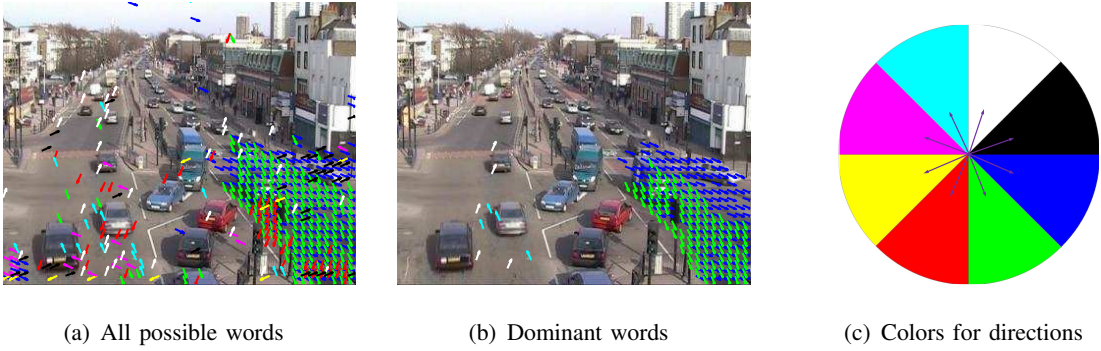

(a) All possible words     (b) Dominant words     (c) Colors for directions

Fig. 4: A comparison between the activity pattern before and after filtering the unnecessary words. The visual words in the left part of image (a) seem chaotic and are filtered out. In (b), the activity is represented better by the selected visual words. The color of the arrow denotes the quantified motion direction, as illustrated in (c).

is the feature vector of clip $t$ given by Eq. (23). Now the training data set $(\mathbf{C}, \mathbf{y})$ is constructed to train the discriminative model- GP. Our task is labeling a new coming video clip $\mathbf{c}^*$ to a traffic state with the highest probability $P(y^*|\mathbf{C}, \mathbf{y}, \mathbf{c}^*)$. For simple illustration the binary classification with two traffic states $y_t \in \{-1, +1\}$ is considered here. The binary classification is easily extended to multiple classification by using the one-against-all or one-against-one strategy.

The general formulation of probability prediction for a new test sample given the training data $(\mathbf{C}, \mathbf{y})$ under a GP model is:

$$p(y^* = +1|\mathbf{C}, \mathbf{y}, \mathbf{c}^*) = \int p(y^*|f^*)p(f^*|\mathbf{C}, \mathbf{y}, \mathbf{c}^*)df^*, \qquad (24)$$

where $p(f^*|\mathbf{C}, \mathbf{y}, \mathbf{c}^*)$ is the distribution of latent variable $f_t$ corresponding to sample $\mathbf{c}^*$. It is obtained by integrating over he latent variable $\mathbf{f} = (f_1, \ldots, f_T)$:

$$p(f^*|\mathbf{C}, \mathbf{y}, \mathbf{c}^*) = \int p(f^*|\mathbf{C}, \mathbf{y}, \mathbf{c}^*, \mathbf{f})p(\mathbf{f}|\mathbf{C}, \mathbf{y})d\mathbf{f} \qquad (25)$$

where $p(\mathbf{f}|\mathbf{C}, \mathbf{y}) = p(\mathbf{f}|\mathbf{y})p(\mathbf{f}|\mathbf{C}) / p(\mathbf{y}|\mathbf{C})$ is the posterior over the latent variables. $p(\mathbf{y}|\mathbf{C})$ is the marginal likelihood (evidence), $p(\mathbf{f}|\mathbf{C})$ is the GP prior over the latent function, which in GP model is a jointly zero mean Gaussian distribution and with the covariance given by the kernel $\mathbf{K}$.

The non-Gaussian likelihood in Eq. (25) makes the integral analytically intractable. We have to resort to either analytical approximation of integrals or Monte Carlo methods. Two well known analytical approximation methods are very suitable for this task, namely the *Laplace* [24] and the *Expectation Propagation* (EP) [25]. They both approximate the non-Gaussian joint posterior as a Gaussian one. In this paper we adopt the *Laplace* method since its computation cost relative lower than EP with comparable accuracy. As introduced in [26] the mean and variance of $f^*$ are obtained as follows:

$$p(f^*|\mathbf{C}, \mathbf{y}, \mathbf{c}^*) = \mathcal{N}(\mu^*, \sigma^*), \tag{26}$$

$$with \quad \mu^* = \mathbf{k}(\mathbf{C}, \mathbf{c}^*)^T \mathbf{K}^- \widetilde{\mathbf{f}}, \tag{27}$$

$$\sigma^{*2} = \mathbf{k}(\mathbf{c}^*, \mathbf{c}^*) - \mathbf{k}(\mathbf{C}, \mathbf{c}^*)^T (\mathbf{K} + \mathbf{W}^-) \mathbf{k}(\mathbf{C}, \mathbf{c}^*), \tag{28}$$

where $\mathbf{W} \triangleq -\nabla\nabla \log p(\mathbf{y}|\widetilde{\mathbf{f}})$ is diagonal. $\mathbf{K}$ denotes a $T \times T$ covariance matrix between $T$ training points. $\mathbf{k}(\mathbf{C}, \mathbf{c}^*)$ is a covariance vector between T training video clips $\mathbf{C}$ and test clip $\mathbf{c}^*$, while $\mathbf{k}(\mathbf{c}^*, \mathbf{c}^*)$ is covariance for test clip $\mathbf{c}^*$, and $\widetilde{\mathbf{f}} = \arg\max_{\mathbf{f}} p(\mathbf{f}|\mathbf{C}, \mathbf{y})$. Given the mean and variance of latent variable $f^*$ for test clip $\mathbf{c}^*$, we compute the prediction probability using Eq. (24).

The covariance function and its hyper-parameters $\Theta$ crucially affect GP models performance. The Gaussian radial basis function (RBF) is one of the most widely used kernels due to its robustness for different types of data and is given as below:

$$K_{RBF}(\mathbf{c}_i, \mathbf{c}_j) = \sigma^2 exp(-\frac{\|\mathbf{c}_i - \mathbf{c}_j\|^2}{2l^2}). \tag{29}$$

$\Theta = [\sigma, l]$ is the hyper-parameter set for RBF. We optimize the hyper-parameters using Conjugate Gradient method [27].

*B. Integration of Transition Information into GP Classifier*

The input video is segmented into clips along time. It can not be ensured that each clip is precise in a traffic state interval. In practice, sometimes the transition of two states occur in a clip, as shown Fig. 5(a). In the other cases, the scene is silent in some clips: there are very few motions, as shown Fig. 5(a). In these two cases, the GP classifier is hard to exactly classify the states. Fortunately, a crowded traffic scene is normally regulated by traffic lights. The transition between two traffic states is rule-based, e.g., the transition from state Fig. 7(a) to state Fig. 7(c) is impossible. The transition information from Sec. IV-B makes significant sense here.

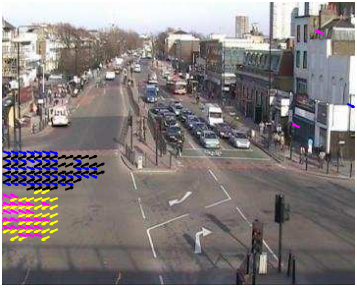 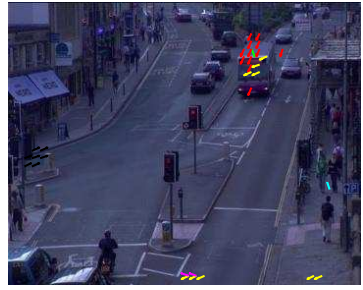

(a) imperfect clip segmentation  (b) too few motions

Fig. 5: Examples of confused traffic states. (a) Imperfect segmented clip may contain motion information belonging to different states. (b) A silent clip contains too few useful motion information. Both of these two cases make the system hard to determine the right state.

We define a state energy for clip $t$ as follows:

$$E(y_t = s_i | y_{t-1} = s_j) = -\log\{p(y_t|\mathbf{c}_t)\} \tag{30}$$

$$+\beta \log\{m_{s_i,s_j}\}(1 - \delta(y_t, y_{t-1}))$$

$$y_t = \arg\min_{y_t = s_i} E(y_t | y_{t-1}) \tag{31}$$

where $p(y_t|\mathbf{c}_t)$ is the likelihood of the $t^{th}$ clip labeled as state $s_i$ given by Eq. (24). $m_{s_i,s_j}$ is the transition probability from state $s_j$ (state of last clip) to $s_i$, and $\delta(y_t, y_{t-1}) = 1$, $if\ y_t = y_{t-1}$, $else\ 0$. $\beta$ is the weight of transition energy and is set experimentally as $0.1$. It means that, if the state does not change, we do not need to care about the transition problem. If the transition of the states happens, we will take the transition information into account and choose the state which has minimal state energy.

## VI. ABNORMAL EVENTS DETECTION

Abnormal events identification is always one of the most interesting and desired capabilities for automated video behavior analysis. However, dangerous or illegal activities often have few examples to learn from and are often subtle. In other words, it is a challenging problem for identifying abnormal events according to their motion patterns for supervised classifier. To tackle this problem, the abnormal events should be defined at first. They are roughly categorized into three groups.

*Rare motions:* The first case is the occurrence of unexpected motions. Such motions do not belong to any typical activities. To detect such abnormal events, in clip $t$ a word set $x'_t$ in size $N'_t$ is defined as the gathering of motions which are not labeled to any learned activity. If $N'_t > th_{word}$, it is confident that some abnormal motions exist during this clip.

*Conflicting Activities:* Second, some activities rarely co-occurred during a clip, i.e. in a specific traffic state, some specific activities rarely occurred. For example, in the state of rightward flow, there should not be any vehicle driving leftward. To detect such abnormal events, we use GP regression to model the temporal relationship among different typical activities during a clip. As we have discussed in Sec. IV-C, the feature vector of clip $t$ is denoted as $\mathbf{c}_t = \{p_{t1}, \cdots, p_{tK}\}$. The value of $p_{ti}$ has underlying relationship with the others. In other words, each value of $\mathbf{c}_t$ can be estimated according to the others in the same clip. Therefore, for each element $p_{ti}$ a GP regression model is constructed. We denote $\mathbf{c}_t^{-p_{ti}} = \{p_{t1}, \cdots, p_{tK}\}$ as the input feature vector of $(K-1)$ dimensions and $p_{ti}$ is the corresponding output value, where $\mathbf{c}_t^{-p_{ti}}$ means that $p_{ti}$ is excluded. A probabilistic prediction about the output value $p_{ti}$ is given by trained GP regression model as:

$$f_*|\mathbf{C}^{-p_i}, \mathbf{p_i}, \mathbf{c}_t^{-p_{ti}} \sim \mathcal{N}(\mu, \sigma), \tag{32}$$

$$\mu = \mathbf{k}_*^T (K - \sigma_n^2 \mathbf{I})^{-1} \mathbf{p_i}, \tag{33}$$

$$\sigma^2 = k(\mathbf{x}_*, \mathbf{x}_*) - \mathbf{k}_*^T (K + \sigma_n^2 \mathbf{I})^{-1} \mathbf{k}_*, \tag{34}$$

where $\mathbf{k}_* = \mathbf{k}(\mathbf{C}^{-p_i}, \mathbf{c}_t^{-p_{ti}})$ and $\mathbf{K} = \mathbf{K}(\mathbf{C}^{-p_i}, \mathbf{C}^{-p_i})$. $f_*$ is the predicted $p_{ti}$ based on the other observed activities. If the observed value $p_{ti}$ is larger than $\mu + 1.96\sigma$, this activity will be vied as conflicting with the others in this clip. $\mu$ is the predicted mean value, $\sigma^2$ is its variance and $(-\infty, \mu + 1.96\sigma)$ is the $97.5\%$ confidence interval. Notice that $p_{ti}$ less than $\mu - 1.96\sigma$ is not viewed as conflict, because in practice an activity causes conflict when its intensity is strong enough. Each activity is modeled by one GP regression model. Therefore, totally $K$ GP regression models are necessary.

*Illegal State Transition:* Finally, a state is followed by another which is forbidden according to the specific traffic rule. Fig. 11 shows an example of an illegal state transition caused by an abnormal event of a fire engine interrupting the current vertical traffic flow and driving rightward. The scene is in vertical flow in $t - 1$ clip and interrupted by fire engine in $t$ clip. During $t + 1$ clip the fire engine is driving cross the scene. Therefore, the $t + 1$ clip would be naturally classified as rightward flow with high probability by GP classifier and the result can be modified by Eq. (31). However, no matter based on our human understanding or the clip's features, this recognition is correct. According to the learned state transition rule as shown in Fig. 7, a rightward flow only follows after the leftward flow. Hence, such case should be determined as an abnormal event. We define a logical judgment to identify such abnormal events. If $p(y_t = s_i | y_{t-1} = s_j) = m(s_i, s_j) < th_{word}$, it will be identified as an illegal state transition, i.e. some abnormal events occur.

*Abnormal Events Localization*: Users are always interesting in the location of of ongoing abnormal events. As discussed in Sec. III, each of visual words contains the position information of its cell in the camera scene. Therefore, all visual words belonging to detected abnormal events can be localized.

We have discussed three kinds of abnormal events and the methods to detect them, respectively. Identifying the abnormal events caused by rare motions and illegal state transition is logic based, which is easy to realize and convenient to apply. [11], [8] use LDA model to estimate the likelihood by iterative sweeps of the Gibbs sampler and detect abnormal events which has low posterior. Different from the methods in [11], [8], for the abnormal events caused by conflicting activities, we use GP regression to model the temporal relationship among activities during a clip. It provides a probabilistic analysis of each activity without complex computation.

## VII. Experiments

### A. Dataset

Experiments were carried out in video data from three complex and crowded traffic scenes regulated by the traffic lights. **QMUL Junction Dataset**: This contains 1 hour of 25 fps video (90000 frames) with frame size $360 \times 288$. The video covers a busy traffic junction containing three major flows in different directions. **QMUL Junction Dataset 2**: This video length is 52 minutes with 25 fps (78000 frames). The frame size is $360 \times 288$. This video is captured in a busy street with particularly busy pedestrian activity. **MIT Dataset** [9]: It consists of 1.5 hour of 30fps (162000 frames) with frame size $720 \times 480$, and captures a far-field traffic scene.

For each dataset, the first 500 video clips (about 25 minute's length) were employed to learn the typical activities and traffic states. The rest of the video sequences were employed to simulate online screened video to test online performance, i.e. 699 clips of QMUL Junction Dataset, 539 clips of QMUL Junction Dataset 2 and 1711 clips of MIT Dataset were used for test.

The ARD kernel was adopted in GP models and the hyper-parameters were optimized by *Conjugate Gradient* [27]. The *Laplace's* approximation method [24] was applied in GP classification models.

To infer the latent variables under the HDP and HDP-HMM, 1000 sweeps of the Gibbs sampler were executed and the first 500 were used as burn-in. To find the best hyper-parameters $(\beta, \alpha)$ for our task, a grid search has been performed on $\beta, \alpha \in \{0.1, 0.5, 1.0, 1.52.0\}$. We analyzed the results with different We got a interesting and useful outcomes: even though the number of clusters increased with larger $\beta$ and $\alpha$, the numbers of typical activities and states always converged when about least $90\%$ of the total motions were explained. These numbers kept consistent when $\beta$ and $\alpha$ were both larger than 0.5. The selected typical activities and states look similar. The additional activities and were generated to explain very rare motions. In this thesis, we are only interested in typical activities and states and we did not use topic models to estimate likelihood or posterior. Therefore, we did not need precise hyper-parameters for the generative models. The hyper-parameters were fixed at $\beta = 2, \alpha = 0.5$ for all experiments. In actual implementation of HDP and HDP-HMM, the hyper-parameters can be optimized by giving a vague gamma prior and sampling them using the scheme proposed in [20].

## B. Learning Typical Activities and States

In QMUL Junction Dataset, the HDP models automatically learned 32 activities in this traffic scene, among which 22 were selected as typical activities (some of them are shown in Fig. 6). Their corresponding percentage computed by Eq. 13 are noted beneath. For a better illustration, all possible motion flows for vehicles and pedestrians are manually painted and marked with alphabetic letters in Fig. 6(q). They are explained as follows:

- Flow a and b: vehicles driving in vertical directions, consists of activities 1, 2, 13, etc.
- Flow c and e: vehicles making a left turn and driving out of the scene. It can be explained by activities 6, 20, and 16 respectively.
- Flow d: vehicles turning left from the left entrance. It is explained by the upper part of activity 4.
- Flow f and g: vehicles making a right turn in the middle of the junction during the vertical flow, shown as activities 9 and 12.
- Flow h (leftward) and i: vehicles driving leftward and part of them making a right turn. It is dominated by activities 3, 17 and 19.

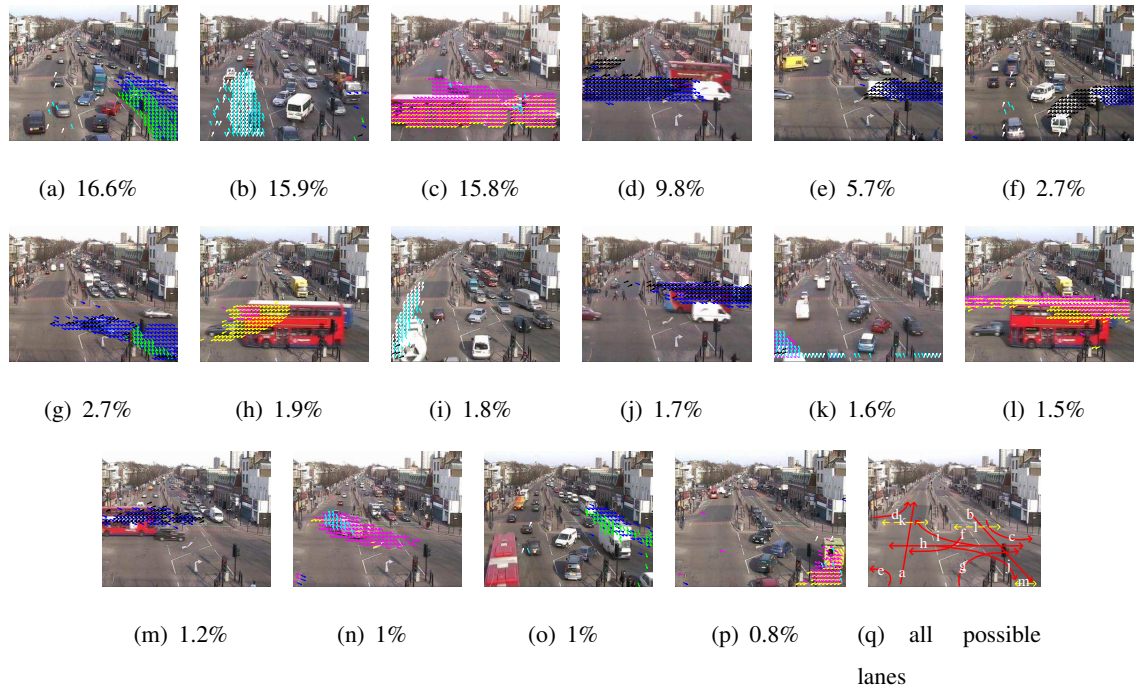

Fig. 6: (a)-(p) Some dominant activities and their percentages discovered by HDP models. (q) Manually labeled legal vehicles driving lanes (red lines) and pedestrians walking lanes (yellow dash lines).

- Flow h (rightward) and j: vehicles driving rightward and part of them making a right turn. It mainly consists of activities 4, 6, 10, 15 and 18.
- Flow k, l and m: pedestrian crossing the road. Activities 15, 17, 18 and 22 show these behaviors.

For QMUL Junction Dataset 2 and MIT Dataset, 21 and 24 typical activities are learned respectively. Due to space constraint, they are not shown and discussed here.

The HDP-HMM automatically learned 9 traffic states. 4 of them are selected as typical states which have the highest percentage among all training clips, as illustrated in Fig. 7(a)-7(d) and their corresponding average feature vectors in the training video shown in Fig. 7(e)-7(h). Fig. 7(i) is the state transition graph with transition probabilities and directions. The are explained as follows

- Vertical flow: Activities 1 and 2 dominate in this interaction. The activities topics such as 5, 8 and 10 related to vertical traffic activities have also relative high values in the histogram.

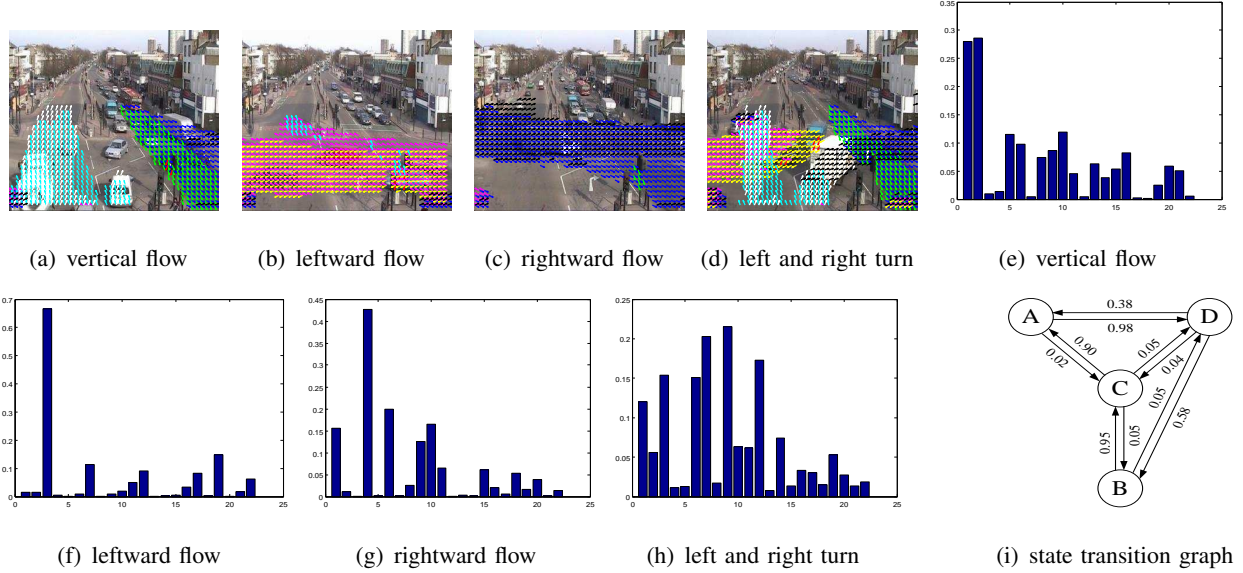

Fig. 7: (a)-(d) are typical traffic states learned by HDP-HMM model and (e)-(h) are their corresponding average components of typical activities. (i) is the state transition graph noted with transition probabilities and directions.

- Leftward flow: It is absolutely dominated by topic 3. Activities 7, 12, 17 and 19 are also important components. The feature values of activities 11, 17 and 22 are relative high because of pedestrians.
- Rightward flow: It mainly consists by activities 4, 6 and 10. Activities 1, 8 and 9 overlap this flow. The feature values of activities 15 and 18 are relative high because of pedestrians.
- Left and right turn: This state happens during the state of vertical flow, when the vertical flow temporally terminates. It is a complicated interaction among a couple of topics, such as activities 1, 3, 6, 7, 8 and 12.

The learned typical traffic states in QMUL Junction Dataset 2 are shown in Fig. 8(a)-8(d) and the states of MIT Dataset are shown in Fig. 8(e)-8(i). QMUL Junction Dataset 2 has two main flows and 4 typical states: vehicles driving vertical without (Fig. 8(c)) or with (Fig. 8(d)) pedestrian; vehicles making a turn at the crossing without (Fig. 8(a)) or with (Fig. 8(b)). The traffic scene in MIT Dataset is relative less busy and interactive than the first QMUL scene: Fig. 8(e) explains a vertical flow. Vehicles from bottom may make a left turn; Fig. 8(e) explains a rightward flow and vehicles making a left turn and driving upward; Fig. 8(g) explains a horizontal flow in two directions. Vehicles may make a left turn in this state; Fig. 8(h) explains vehicles driving downward from top and pedestrian crossing the road; Fig. 8(i) illustrates that, vehicles stop behind the crosswalk and pedestrian cross the road.

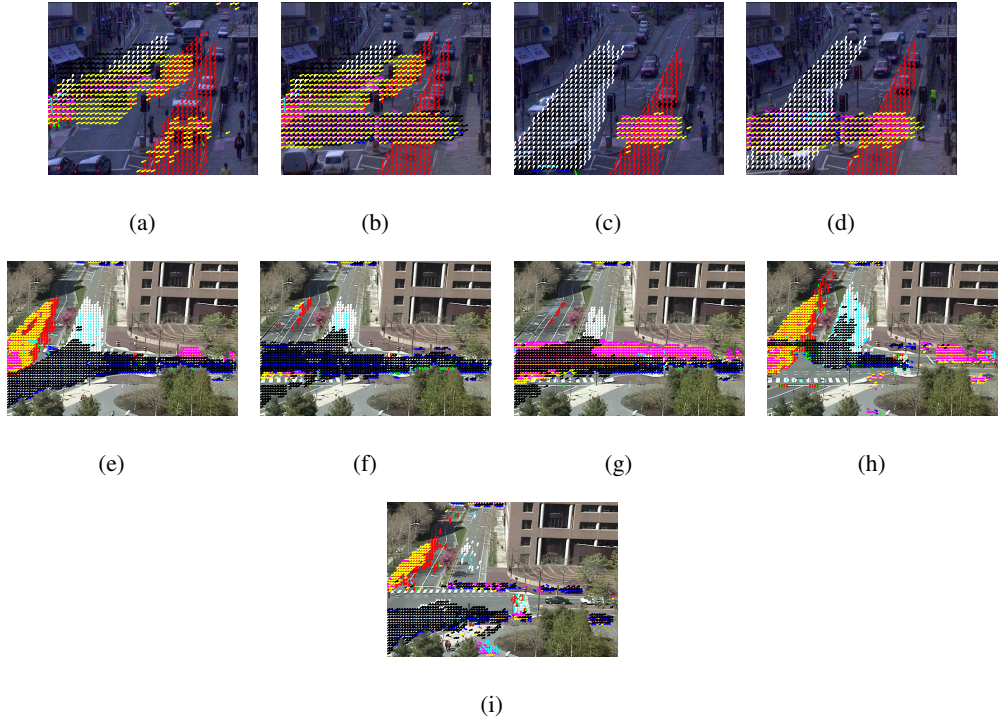

Fig. 8: Typical traffic states learned by HDP-HMM model for QMUL Junction Dataset 2 (a)-(d) and MIT Dataset (e)-(i).

## C. Traffic States Recognition

The GP classifier was firstly trained with learned activities and states. The screened video sequence was segmented into clips of 75 frames each.

Our experimental results are compared with the other popular methods: Dual-HDP model [9], Markov Clustering Topic Models(MCTM) [8], LDA and HMM. They adopted diverse length of video clip ranging from 1 second to 10 seconds. The experimental results are directly cited from [19] (for QMUL Dataset) and [9](for MIT Dataset). From the comparison in Tab. I we see that our model outperforms other three popular methods in terms of classification results in the QMUL Dataset. In contrast to the Dual-HDP model in the MIT Dataset as listed in Tab. II, our methods also achieved better classification results. Furthermore, Dual-HDP model is a batch processing. To validate our method, we have executed one more experiment in the QMUL Junction Dataset 2. The results is listed in Tab. III.

| State | MCTM | | | | LDA | | | | HMM | | | | Ours | | | |
|---|---|---|---|---|---|---|---|---|---|---|---|---|---|---|---|---|
| | L | R | V | VT | L | R | V | VT | L | R | V | VT | L | R | V | VT |
| Left | **.99** | .00 | .00 | .01 | .49 | .44 | .00 | .06 | **.98** | .00 | .01 | .01 | **1.0** | .00 | .00 | .00 |
| Right | .00 | **.94** | .01 | .05 | .00 | **1.0** | .00 | .00 | .00 | **.92** | .08 | .00 | .00 | **.99** | .00 | .01 |
| Vertical | .00 | .00 | **.77** | .22 | .01 | .17 | **.82** | .00 | .02 | .01 | **.69** | .28 | .00 | .00 | **.98** | .00 |
| Vertical-Turn | .31 | .05 | .20 | **.43** | .01 | .21 | .30 | **.46** | .49 | .04 | .32 | **.15** | .05 | .00 | .00 | **.95** |
| Average Accuracy | .78 | | | | .69 | | | | .69 | | | | .98 | | | |

TABLE I: Comparison of Classification results between our methods and others popular methods for QMUL Juction Dataset: The results of MCTM, LDA and HMM are cited from [19].

| | State | Dural-HDP | | | | | Ours | | | | |
|---|---|---|---|---|---|---|---|---|---|---|---|
| | | a | b | c | d | e | a | b | c | d | e |
| Manually label | a | 149 | 0 | 2 | 0 | 0 | 610 | 4 | 5 | 0 | 3 |
| | b | 8 | 74 | 4 | 2 | 11 | 3 | 402 | 0 | 2 | 0 |
| | c | 10 | 3 | 60 | 1 | 2 | 3 | 2 | 304 | 2 | 0 |
| | d | 4 | 0 | 2 | 88 | 11 | 7 | 8 | 10 | 222 | 0 |
| | e | 4 | 2 | 6 | 5 | 92 | 6 | 5 | 4 | 8 | 102 |

TABLE II: Classification performance for the MIT Dataset.

It is worth point out that some clips were falsely recognized by traditional GP classifier and corrected by our model. For example, it is ambiguous to determine whether the state in Fig. 9 belongs to state Fig. 8(e) or Fig. 8(f) only based on its appearance. It was falsely classified as the second one with higher probability by GP classifier. Because its previous clip is in the state as Fig. 8(e), it is successfully corrected by using transition information, as described in Sec. V-B.

|   | Our Classification | | | |
|---|---|---|---|---|
|   | a | b | c | d |
| a | 86 | 2 | 1 | 2 |
| b | 2 | 264 | 0 | 4 |
| c | 0 | 0 | 188 | 2 |
| d | 0 | 2 | 0 | 76 |

TABLE III: Classification performance for QMUL dataset 2.

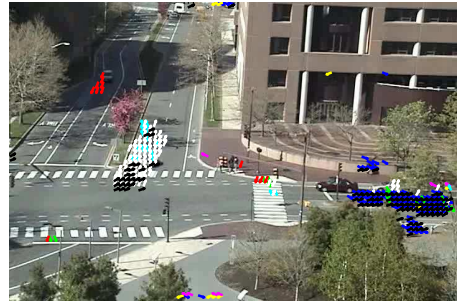

Fig. 9: Example of falsely classified by GP classifier.

### D. Anomaly Detection

Then the proposed framework's performance of detecting the abnormal events defined in Sec. VI is evaluated in each dataset. In the scene of QMUL Junction Dataset, the main abnormal events include Jaywalking, illegal U-turn and emergencies caused by ambulances, fire engines and police cars. Fig. 10 illustrates two detected abnormal events caused by rarely occurring motions in the QMUL Junction scene. For instance, the ambulance is driving in an absolutely forbidden direction in the lane, whose motions have never occurred in the training data (Fig. 10(a)).

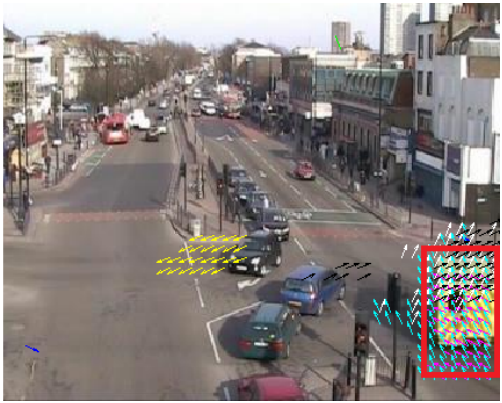
(a) Ambulance

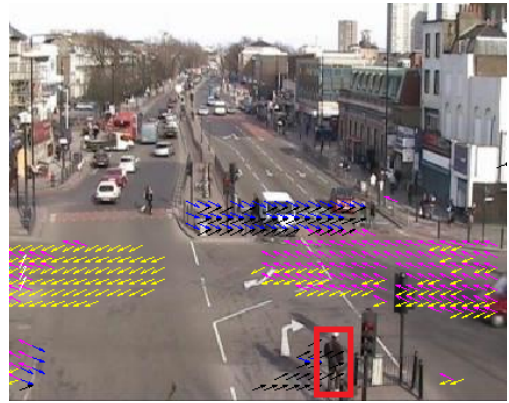
(b) Pedestrian walking in improper area

Fig. 10: Examples of abnormal events caused by rarely occurring motions. In the training dataset such motions have rarely or never occurred. They do not belong to any typical activities. The red boxes highlight the abnormal agents.

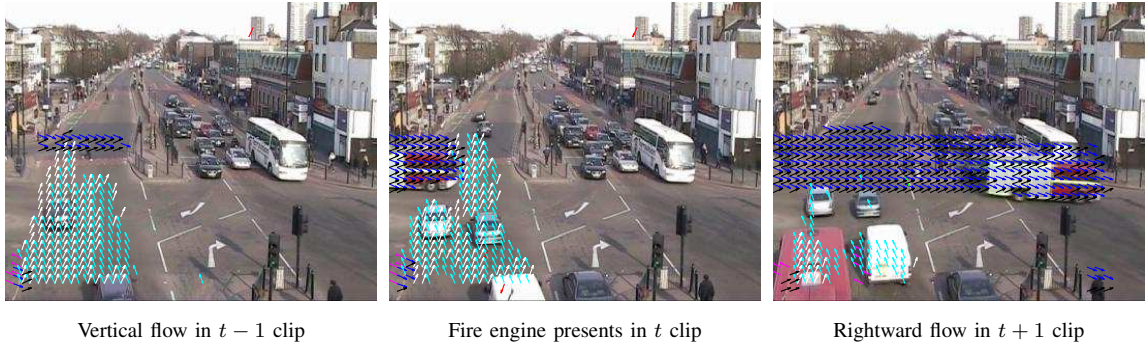

Vertical flow in $t-1$ clip | Fire engine presents in $t$ clip | Rightward flow in $t+1$ clip

Fig. 11: Example of abnormal event caused by illegal state transition. A fire engine interrupts the current vertical flow. The red boxes highlight the abnormal agents.

In Fig. 11, the traffic state is forced to change in an illegal ordering due to the fire engine. The rightward flow should not follow the vertical flow according to the learning results by HDP-HMM models. Therefore, the clip was identified as abnormal event, even though its appearance is definitely a right flow state.

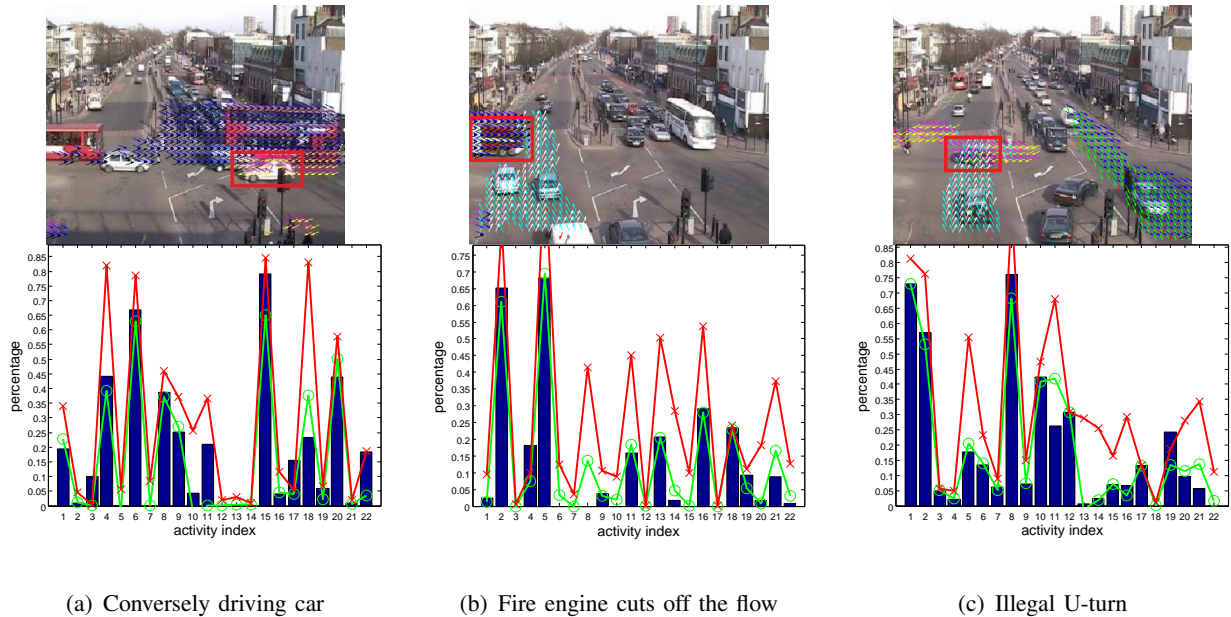

(a) Conversely driving car    (b) Fire engine cuts off the flow    (c) Illegal U-turn

Fig. 12: The first row shows three examples of detected abnormal events caused by conflicting activities. The second row illustrate the observed values (blue bars) of each activity in given scene respectively. The green curves (circles) note the mean predicted values $m$ and the red curves (crosses) equal $m + 1.96\sigma$, i.e. the upper bounds of $97.5\%$ confidence region.

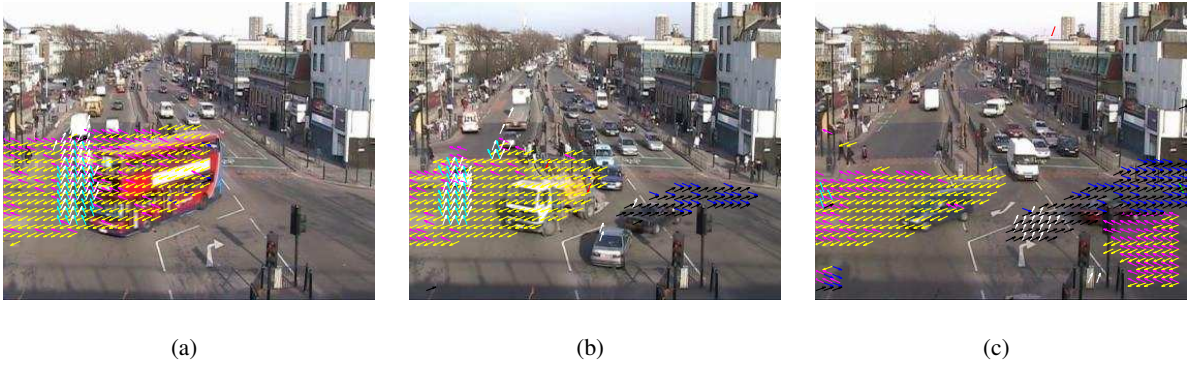

Fig. 13: Falsely detected Abnormal events by the GP regression models.

Normally, abnormal events are caused by conflicting activities like Jaywalking, illegal U-turn converse driving or aggressively cutting in other roads. We illustrate an example for each type of such conflicting activities in Fig. 12. The conflicting activities were detected by our GP regression models as discussed in Sec. VI. If the observed value (blue bar) of any activities is larger than its predicted value (green curve and circle) as $1.96\sigma$ (red curve and cross), it is judged as a conflict activity against the others. The abnormally acting agents are marked by red boxes. They are analyzed in detail as following (all the related atomic activities are founded in Fig. 6):

- Fig. 12(a) shows a police car driving conversely. This counter flow induces the value of activities 3 and 17 in the former clip, activities 3 and 19 in the latter clip.
- A fire engine cuts off the vertical flow (see Fig. 12(b)) and causes the activity 4 much stronger than the prediction.
- In Fig. 12(c) the activity 19 is abnormal because of a vehicle making an illegal U-turn.

Some falsely detected abnormalities are shown in Fig. 13. The red double-decker bus is detected as a U-turn agent due to its big size. In Fig. 13(c), a conflict activity is detected occurring in the right bottom of the camera scene, because in this state, there should not be a leftward traffic flow. However, this alarm is a misunderstanding by our models because of the bad video clip segmentation. This clip contains the state transition and is classified as the left and right turn state. Therefore, the GP regression models thought the activity conflicting with others in this state and judged it as an abnormal event. Actually, this activity occurred when the state has already changed into the state of leftward flow.

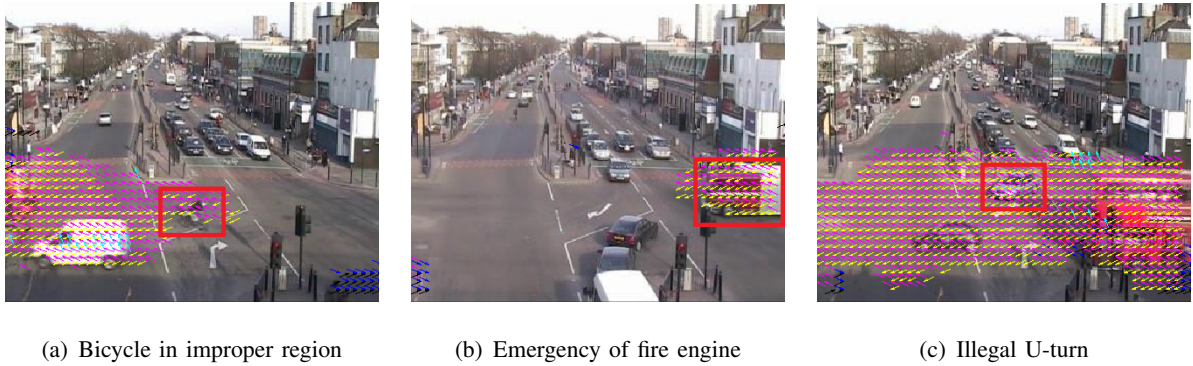

(a) Bicycle in improper region     (b) Emergency of fire engine     (c) Illegal U-turn

Fig. 14: Examples of missing detected abnormal events.

Fig. 14 shows some missing detected abnormal events. Because our method is beyond detecting, the categories of activity agents are not considered. For example, if a pedestrian is walking along the path of vehicles, it will not be detected as an abnormality, as shown in Fig. 14(a). In Fig. 14(b), before the fire engine drives into the camera scene, all vehicles have stopped and wait for its pass. Therefore, there is no activity in conflict with the fire engine. The scene is classified as leftward state. Because of its previous state is the sate left and right turn, this transition is legal. That is why this emergency was undetected. A car is making an illegal U-turn in Fig. 14(c). However, its activity seems identical with others in the leftward state. Hence, it is also not identified as an abnormal activity. The detection and tracking based approaches would perform better in this case.

We provide a manually interpreted summary of the categories of abnormal events of each dataset in Tab. IV. Notice that, each entire abnormal event is counted as one event, no matter how many clips it spans. The false detection means that, a clip is detected as an abnormal clip, but there is not any abnormal event of interest. The overall false positive rates is defined as:

$$FPR = \frac{\text{Number of falsely detected clips}}{\text{Number of test clips}}. \tag{35}$$

From the summary of experimental results we can see that, our method successfully detected most of the abnormal traffic events while causes low overall false positive rates in the three benchmark datasets. However, it seems weak in detecting "improper region" because the proposed method is beyond object detection. In other words, it is the abnormal motions of any agent in specific case cause the anomaly alarm rather than the category of agent. A concrete example is given in Fig. 14(a). Moreover, in the experiments we find that, our trained model own the ability of working in real time beyond the computation bottleneck of optical flow.

| Dataset | Results | Jaywalking | Emergency | Illegal turning | Near collision | Strange driving | Improper region | False detection | Overall TPR | Overall FPR |
|---|---|---|---|---|---|---|---|---|---|---|
| QMUL junction | GT | 19 | 4 | 10 | 2 | 1 | 2 | \ | 66% | 2.6% |
| | Ours | 11 | 3 | 7 | 2 | 2 | 0 | 18 | | |
| QMUL junction 2 | GT | 21 | \ | \ | 2 | 4 | \ | \ | 63% | 2.1% |
| | Ours | 14 | \ | \ | 2 | 1 | \ | 7 | | |
| MIT | GT | 14 | \ | 34 | \ | 1 | 13 | \ | 65.7% | 2.9% |
| | Ours | 7 | \ | 28 | \ | 1 | 5 | 43 | | |

TABLE IV: Summary of discovered abnormal events in different datasets. Overall true positive and false positive rates are also given. The "\" symbol indicates that there is not such event in the dataset. "Gt", "TPR" and "FPR" mean ground truth, truth positive rate and false positive rate respectively.

## VIII. CONCLUSIONS

In this paper, a novel unsupervised learning framework has been proposed to model the activities and interactions, to recognize global interactions and to identify abnormal events in crowded and complicated traffic scenes. Through combining the advantages of both generative models (HDP models) and discriminative ones (GP models), the formulated approach provides an effective solution to the problems of high-level video events recognition and abnormal events detection. First, owing to its computation efficiency as well as comparative reliability in the far-field surveillance data, the quantized optical flow is adopted in this work as the low-level motion features. Then, a non-parametric generative HDP model is utilized to analyze the input video and learn the main activities and interactions in a unsupervised way. Next, each of the learned activities and interactions are represented as a combination of the local motions and the combination of activities respectively. Finally, each activity and interaction, a GP model is trained using the aforementioned representation for classification tasks and anomaly detection. The experimental results demonstrate that the approach outperforms other popular approaches in both classification accuracy and computation efficiency. In particular, the improved GP classifier is capable to correct the falsely-classified clips by the original GP classifier. There are many exciting avenues for future research. First, it will be interesting to incorporate the segmentation methods [28] into our proposed framework. Second, we will test the proposed algorithm using high-resolution remote sensing images, where the visual features are clear and informative [29], [30]. Third, we would also like to compare the performance of our model to the recent CNN model [31]. Finally, the current model only takes into account the simple temporal dependencies within a clip in detecting conflicting activities. It could result in poor performances of abnormality detection in scenes, of which the traffic state is quite obscure because of the absence of traffic lights. One possible solution is to use additional GIS data to enhance the classification task and anomaly detection.

## ACKNOWLEDGMENT

This research is funded by German Research Foundation DFG within Priority Research Programme 1894 "Volunteered Geographic Information: Interpretation, Visualisation and Social Computing". The authors gratefully acknowledge the support.